\documentclass{article}

\usepackage[final]{corl_2026} 

\usepackage[utf8]{inputenc} 
\usepackage[T1]{fontenc}    
\usepackage{hyperref}       
\usepackage{url}            
\usepackage{booktabs}       
\usepackage{arydshln}
\usepackage{nicefrac}       
\usepackage{microtype}      
\usepackage{amsfonts}       
\usepackage{etoc}

\usepackage{amsmath}
\usepackage{amssymb}
\usepackage{mathtools}
\usepackage{amsthm}
\usepackage{graphicx}
\usepackage{subcaption}
\usepackage{enumitem}
\usepackage{makecell}
\usepackage{algorithm}
\usepackage{algorithmic}
\usepackage{booktabs}
\usepackage{tabularx}
\usepackage{graphicx}

\newcommand{\ie}{\textit{i.e.}}
\newcommand{\eg}{\textit{e.g.}}
\newcommand{\ours}{PGPG}
\newcommand{\fullours}{\textbf{P}rogress-\textbf{G}uided \textbf{P}reference \textbf{G}eneration}

\newcommand{\stdv}[1]{\scalebox{0.7}{\textcolor{gray}{$\pm$#1}}}

\newcommand{\state}{\mathbf{s}}

\newcommand{\observation}{\mathbf{o}}
\newcommand{\transition}{\mathcal{T}}
\newcommand{\action}{\mathbf{a}}

\newcommand{\obsspace}{\mathcal{O}}
\newcommand{\statespace}{\mathcal{S}}
\newcommand{\actionspace}{\mathcal{A}}

\newcommand{\rewardfunc}{r}
\newcommand{\reward}{r}

\newcommand{\expertdataset}{\mathcal{D}_{\text{demo}}}

\newcommand{\prefdataset}{\mathcal{D}_{\text{pref}}}
\newcommand{\rolloutdataset}{\mathcal{D}_{\text{rollout}}}

\title{Progress-Guided Preference Generation for Aligning Generative Control Policies}

\author{
\parbox{\textwidth}{\centering
\vspace{0.2in}
Tung M. Luu\textsuperscript{1}\quad
Tri Ton\textsuperscript{1}\quad
Duc M. Nguyen\textsuperscript{2,3}\quad
Xuan Thanh Nguyen\textsuperscript{1}\quad
Nguyen Gia Binh\textsuperscript{2,3}\quad \\[0.05in]
Quang Tan Nguyen\textsuperscript{3}\quad
Vu Duong\textsuperscript{2}\quad
Daniel Sonntag\textsuperscript{6,7}\quad
Duy M. H. Nguyen\textsuperscript{4,5,6}\quad \\[0.05in]
Vien Anh Ngo\textsuperscript{2,3}\quad 
Chang D. Yoo\textsuperscript{1}\\[1.2em]
    \footnotesize\normalfont
    \textsuperscript{1}KAIST
\quad
\textsuperscript{2}Center for AI Research \& CECS, VinUniversity
\quad
\textsuperscript{3}VinRobotics
\quad
\textsuperscript{4}University of Stuttgart
\\[0.05in]
\textsuperscript{5}Max Planck Research School for Intelligent Systems
\quad
\textsuperscript{6}DFKI
\quad
\textsuperscript{7}University of Oldenburg
}
}
\begin{document}
\maketitle


\begin{abstract}
    Generative policies have achieved strong performance in imitation learning tasks, but their success is often constrained by the limited availability of expert demonstrations. Although preference-based fine-tuning can leverage suboptimal rollouts to improve performance, existing approaches heavily rely on expensive human annotations. In this work, we introduce \fullours{} (\ours{}), an unsupervised preference alignment framework that eliminates the need for human-in-the-loop supervision. Our method leverages ordered coverage alignment to measure sequential progress relative to a small set of expert demonstrations, enabling the automatic extraction of pseudo-preference labels from autonomously collected rollouts. Furthermore, we introduce key enhancements to improve the reliability of these progress-guided preferences, enabling effective fine-tuning of pretrained generative policies without requiring manually designed reward functions. Extensive experiments across a suite of challenging long-horizon manipulation tasks demonstrate that \ours{} outperforms standard supervised fine-tuning baselines. Moreover, our conceptually simple framework is broadly applicable across multiple classes of generative policies, demonstrating strong versatility when integrated with both diffusion and flow-matching policies. Project site is available at: \url{https://pgpg2026.github.io}.
\end{abstract}



\section{Introduction}
\vspace{-0.1in}
Learning robotic policies that generalize beyond curated demonstrations remains a central challenge in reinforcement learning and imitation learning \cite{christiano2017deep,hejna2024contrastive,chen2022offline}. Behavior cloning can produce strong initial policies from expert data, but it is often brittle under distribution shift and highly sensitive to data quality. This issue is particularly pronounced in real-world settings, where expert trajectories are scarce, training coverage is limited, and the available demonstrations may themselves be suboptimal \cite{mandlekar2021matters,chi2023diffusion}. Reinforcement learning, by contrast, can improve robustness through trial-and-error interaction. However, it typically requires carefully engineered reward functions, extensive environment interactions, and challenging exploration processes, all of which may be inefficient or unsafe in real-world control scenarios \cite{clark2016faulty,christiano2017deep,wang2022diffusion}. Recently, preference-based learning \cite{casper2023open} has emerged as a promising middle ground between imitation learning and reinforcement learning \cite{zhang2024grape,hejna2024contrastive}. Instead of relying solely on expert demonstrations or manually engineered reward functions, preference-based methods replace explicit reward design with relative comparisons between trajectory segments. This allows policies to learn from simple preference queries rather than handcrafted scalar rewards \cite{christiano2017deep,lee2021pebble,lee2021b,hejna2024contrastive}. This paradigm is particularly attractive for robotics, where designing reliable reward functions is difficult, and where improving the performance and robustness of pretrained policies while minimizing costly environment interaction is highly desirable \cite{zhang2024grape,choi2024listwise,mu2025clarify,miao2025uncertainty,cai2025predictive}.

A major obstacle to scaling preference-based learning is its heavy reliance on human feedback, which is inherently costly and labor-intensive \cite{hejna2023inverse,hejna2024contrastive}. This bottleneck is particularly restrictive for modern generative control policies, such as diffusion \cite{chi2023diffusion} and flow-matching models \cite{zhang2025flowpolicy}. While these architectures are highly expressive and excel at capturing multimodal action distributions, they currently lack scalable alignment mechanisms when labeled preferences are scarce \cite{miao2025uncertainty}. Because existing alignment methods typically assume access to abundant, high-quality human annotations, their real-world applicability remains severely limited \cite{zhang2024grape,xia2025robotic,miao2025uncertainty}. These limitations highlight an open question: \textit{How can we automatically extract meaningful preference signals from autonomously collected rollouts to effectively fine-tune generative policies without human-in-the-loop supervision?}

To this end, we propose \fullours{} (\ours{}), an unsupervised framework for fine-tuning generative control policies. The key idea is to use ordered coverage alignment to measure the continuous progress of collected rollouts relative to a small set of expert demonstrations, and then convert this progress into preference labels for segment pairs. Importantly, the resulting signal does not merely evaluate whether a rollout reaches the final goal, but also whether it exhibits ordered progress that is structurally consistent with expert behavior. To make this signal suitable for preference ranking, we transform cumulative progress into marginal rewards, avoiding temporal double-counting and enabling meaningful comparisons between trajectory segments. We further aggregate progress across multiple demonstrations, which is particularly important in robotic manipulation tasks where multiple valid strategies may exist and a single demonstration should not overly constrain learning. In this way, \ours{} transforms limited demonstration coverage into a scalable source of supervision for preference alignment. Extensive experiments across a suite of challenging manipulation tasks from Robomimic \cite{mandlekar2021matters} demonstrate that \ours{} can effectively fine-tune generative policies using only a small set of expert demonstrations together with a large collection of autonomously collected rollouts. We further demonstrate the versatility of our framework by seamlessly integrating it with multiple classes of generative policies, including Diffusion Policy \cite{chi2023diffusion} and Rectified Flow Policy \cite{zhang2025reinflow}.

\section{Related Work}
\vspace{-0.05in}
\paragraph{Preference-based Reinforcement Learning.} Extensive research has explored aligning agent behavior through pairwise trajectory ranking \cite{christiano2017deep,ibarz2018reward,biyik2020active,lee2021pebble,lee2021b,mu2025clarify,choi2024listwise}. The traditional paradigm in preference-based reinforcement learning (PbRL) typically operates in two stages: first fitting a proxy reward model from preference data, and subsequently using the learned reward to guide downstream policy optimization \cite{christiano2017deep,lee2021b,kim2023preference,choi2024listwise,luu2026video}. While this paradigm has demonstrated strong performance across domains ranging from Atari \cite{christiano2017deep,ibarz2018reward} to large language models \cite{ouyang2022training}, the intermediate reward-modeling stage introduces additional optimization complexity. More recently, methods such as Direct Preference Optimization (DPO) \cite{rafailov2023direct} and Contrastive Preference Learning (CPL) \cite{hejna2024contrastive} have eliminated the need for an explicit reward model. Instead, these approaches directly fine-tune policies to satisfy preference constraints through a simplified classification objective. Owing to their simplicity and efficiency, these methods have been adopted for fine-tuning generative control policies \cite{zhang2024grape,miao2025uncertainty,shan2025forward}.

However, applying PbRL to robotic control remains heavily bottlenecked by the substantial human effort required to label large amounts of interaction data \cite{hejna2023inverse,an2023direct,yuan2024uni,wang2024rl,luu2025enhancing,luu2025policy}. To eliminate this human-in-the-loop burden, our framework automatically extracts preference labels from a limited set of offline expert demonstrations. Furthermore, while recent automated labeling pipelines rely on large language models to generate Python reward functions \cite{zhang2024grape}, \ours{} derives preference labels directly from expert demonstrations, naturally grounding the learning signal in the visual and kinematic structure of the task.

\paragraph{Generative Control and Preference Alignment.}
Recent advances in generative modeling, particularly diffusion and flow-matching frameworks \cite{ho2020denoising,lipman2022flow,pham2025mdsgen,ton2025taro}, have demonstrated a remarkable capability to capture complex, multi-modal distributions. This high expressiveness makes them uniquely suited for decision-making and robotic control, triggering broad adoption across offline RL \cite{janner2022planning,ajay2023conditional}, online RL \cite{yang2023policy,ren2025diffusion,zhang2025sac}, and manipulation tasks \cite{chi2023diffusion,sridhar2024nomad,ze20243d}. By framing policies as generative processes, these approaches model multi-modal action trajectories more robustly than conventional Gaussian or VAE-based baselines \cite{wang2022diffusion,nguyen2026fast}.
Early foundational works like DQL \cite{wang2022diffusion} integrated diffusion processes directly into Q-learning frameworks. Subsequent iterations focused on stabilizing optimization and improving evaluation. For instance, SfBC \cite{chen2022offline} decoupled behavior cloning from explicit action selection, while CEP \cite{lu2023contrastive} introduced generalized energy-guided sampling. To address the computational overhead of iterative inference denoising, recent lightweight formulations such as FQL \cite{fql_park2025} and OFQL \cite{nguyen2026onestep} drastically minimize sampling complexity while preserving competitive policy performance.

Concurrently, generative control models have emerged as competitive backbones for preference-based alignment. While early methods like AlignDiff \cite{dong2024aligndiff} and FTB \cite{zhang2024flow} utilize complex, two-stage diffusion planning loops, recent frameworks like FKPD \cite{shan2025forward} and UAPA \cite{miao2025uncertainty} pivot toward highly efficient, one-step direct policy optimization. While \ours{} operates within this efficient one-step alignment paradigm, it bypasses the core bottleneck of existing literature. Rather than relying on costly, large-scale human annotations, \ours{} synthesizes reliable preference labels directly from autonomously collected rollouts by anchoring progress to a small expert dataset.

\section{Preliminaries}
\vspace{-0.05in}
\paragraph{Preference-based Reinforcement Learning.} Reinforcement learning (RL) algorithms \cite{sutton2018reinforcement} typically consider a Markov decision process (MDP), formally defined as a tuple $\mathcal{M}=(\statespace,\actionspace,\transition,\rewardfunc, \gamma, p_0)$, where $\statespace$ and $\actionspace$ denote the state and action spaces, $\transition(\state'|\state,\action)$ and $\rewardfunc(\state,\action)$ represent the transition dynamics and reward function, $\gamma \in [0,1)$ is the discount factor, and $p_0$ denotes the initial state distribution. In our setting, the agent additionally receives a visual observation $\observation \in \obsspace$, represented as an image rendered from the underlying state $\state$. In many real-world applications, the reward function is not directly available, resulting in a reward-free MDP $\mathcal{M}\setminus\rewardfunc$. Preference-based RL \cite{christiano2017deep} addresses this challenge by learning a reward function $\rewardfunc_{\phi}$ from preference data, typically provided by human annotators. Given an unlabeled dataset of trajectory segments, annotators compare sampled pairs and indicate which segment they prefer. These pairwise comparisons form a preference dataset $\prefdataset = \{(\sigma^w, \sigma^l)\}$, where each trajectory segment of length $L$ is defined as $\sigma = (\state_1, \action_1, \dots, \state_L, \action_L)$, and $\sigma^w$ is preferred over $\sigma^l$. Recent methods \cite{christiano2017deep,ibarz2018reward} typically employ the Bradley-Terry model \cite{bradley1952rank} to learn the reward function $\rewardfunc_{\phi}$ from these preference comparisons.

\paragraph{Improving Generative Policy via Preference Alignment.} In this work, we consider a family of generative control policies, including diffusion policies \cite{chi2023diffusion} and flow-matching policies \cite{zhang2025flowpolicy}. Let $\rolloutdataset = \{\tau\}$ denote the dataset of additional rollouts collected by a pretrained policy interacting with the environment, where each trajectory is defined as $\tau=(\state_1, \action_1, \dots, \state_H,\action_H)$. Given a small set of expert demonstrations $\expertdataset$ and a large collection of autonomously collected rollouts $\rolloutdataset$, our goal is to effectively fine-tune the pretrained policy from both data sources. 

Because explicit environmental rewards are unavailable, directly applying RL algorithms such as PPO \cite{schulman2017proximal} is challenging. We therefore formulate the fine-tuning stage as a preference alignment problem, leveraging the limited set of expert demonstrations as a weak supervisory signal to extract preferences from the uncurated rollouts. We follow recent work \cite{zhang2024grape,shan2025forward,miao2025uncertainty} that adopts contrastive preference learning (CPL) \cite{hejna2024contrastive} to align generative policies with human preferences. Concretely, for a pair of preferred and rejected trajectory segments $(\sigma^w, \sigma^l)$, the CPL objective is:
\begin{equation}\label{eq:cpl_loss}
    \mathcal{L}_{\text{CPL}}^{(\sigma^w,\sigma^l)}(\theta) = -\log\sigma\left(\alpha \cdot [\sum_{t=1}^L\gamma^t\log\frac{\pi_{\theta}(\action_t^w|\state_t^w)}{\pi_{\text{ref}}(\action_t^w|\state_t^w)} - \sum_{t=1}^L\gamma^t\log\frac{\pi_{\theta}(\action_t^l|\state_t^l)}{\pi_{\text{ref}}(\action_t^l|\state_t^l)}]\right),
\end{equation}
where $\sigma(\cdot)$ is the Sigmoid function, $\sigma^w = (\state_t^w, \action_t^w)_{t=1}^L$ and $\sigma^l = (\state_t^l, \action_t^l)_{t=1}^L$ denote the preferred and rejected trajectory segments, respectively, $\pi_{\text{ref}}$ denotes the reference policy, typically instantiated as the pretrained policy, and $\alpha > 0$ controls the deviation from the reference policy. Detailed learning objectives for diffusion and flow-matching policies are provided in the Appendix.

\paragraph{Sequence Matching.} Sequence-matching problems \cite{garreau2014metric} have been widely explored in inverse RL to align a learner's trajectories with expert demonstrations. While many approaches rely on optimal transport \cite{haldar2022watch,luo2023optimal,fu2024robot} to match expert trajectory distributions, these methods often struggle to preserve temporal consistency and are prone to goal abandonment in long-horizon robotic tasks \cite{huey2025imitation}. This limitation motivates the need for a structurally ordered metric. In this work, we adopt ordered coverage alignment (ORCA) \cite{huey2025imitation} to quantify the continuous progress of collected rollouts relative to expert demonstrations. This measure is particularly well-suited for our framework because it inherently handles temporal misalignments, such as variations in execution speed between the learner and the expert. Concretely, given a learner trajectory $\tau$ of length $H$ and an expert demonstration $\tilde{\tau}$ of length $\tilde{H}$, the ORCA step-wise progress at timestep $t$ in the learner's trajectory is computed as:
\begin{equation}\label{eq:orca_reward}
\mathcal{R}(t, \tau, \tilde{\tau}) = C_{t,\tilde{H}-1}P_{t,\tilde{H}},
\end{equation}
where $P_{t,j} = \exp(-\lambda\|\observation_t - \tilde{\observation}_j\|)$ denotes the probability that the learner occupies the $j$-th demonstration subgoal at timestep $t$, and $\lambda$ is a temperature hyperparameter. The term $C_{t,j}$ represents the recursive historical coverage, computed via dynamic programming as $C_{t,j} = \max\{C_{t-1, j}, C_{t, j-1}P_{t,j}\}$. This formulation tracks the probability that the learner has successfully completed all preceding subgoals up to $j$ in the correct order. Intuitively, Eq.~(\ref{eq:orca_reward}) assigns a high score only when the agent reaches and remains at the final target without skipping intermediate subgoals. Additionally, the ordered coverage is non-decreasing over time: once the learner has covered a subgoal at a given timestep, it remains at least equally successful at future timesteps. In this work, we leverage this continuous progress measure to automatically derive preference labels for segment pairs, enabling preference learning without a human-in-the-loop.

\begin{figure*}[t]
\begin{center}
    \includegraphics[width=1.0\linewidth]{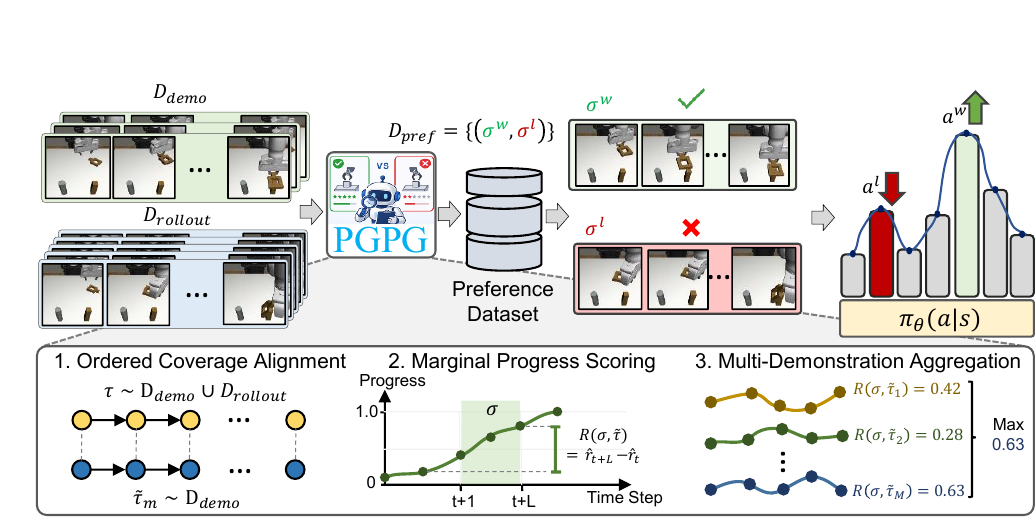}
\end{center}
\vspace{-0.05in}
\caption{\textbf{Overview of \ours{}}. Autonomously collected rollouts ($\mathcal{D}_{\text{rollout}}$) are scored against expert references ($\mathcal{D}_{\text{demo}}$) using an alignment pipeline consisting of Ordered Coverage Alignment, Marginal Progress Scoring, and Multi-Demonstration Aggregation. The aggregated segment scores are then used to derive the preference labels. The resulting dataset $\mathcal{D}_{\text{pref}}$ provides structured, dense pseudo-preferences to stably fine-tune generative policies $\pi_\theta(a|s)$ without manual reward engineering.}
\vskip -0.1in
\label{fig:PGPG_overview}
\end{figure*}

\section{Method}
\vspace{-0.1in}
While the preference learning objective in Eq.~(1) provides a powerful mechanism for aligning generative policies, existing approaches typically rely on human-provided preference labels. This manual annotation process is expensive, difficult to scale, and introduces a major bottleneck for robotic control tasks. To eliminate this dependence on human annotation, we propose \fullours{} (\ours{}), a framework that automatically generates progress-guided preferences to fine-tune policies toward expert behavior, as shown in Figure~\ref{fig:PGPG_overview}.

\subsection{Progress-Guided Preference Generation}

\paragraph{Marginal Progress Scoring.} Let $\mathcal{D} = \expertdataset \cup \rolloutdataset$ denote the combined dataset used to fine-tune the pretrained policy. For each trajectory $\tau \sim \mathcal{D}$, we first compute the progress score at every timestep $t$ using the ORCA metric anchored against a reference demonstration $\tilde{\tau} \sim \expertdataset$. This yields a progress-labeled trajectory $\hat{\tau} = (\state_1, \action_1, \hat{\reward}_1, \dots, \state_H, \action_H, \hat{\reward}_H)$, where $\hat{\reward}_t = \mathcal{R}(t, \tau, \tilde{\tau})$, resulting in a scored dataset $\hat{\mathcal{D}}=\{\hat{\tau}\}$. To construct the preference dataset, a naive approach would sample two trajectory segments $\sigma_1$ and $\sigma_2$ from $\hat{\mathcal{D}}$, calculate the sum of their step-wise scores $\sum_{t \in \sigma_1} \hat{\reward}_t$ and $\sum_{t \in \sigma_2} \hat{\reward}_t$, and label the segment with the higher sum as preferred. However, applying this naive summation directly to the raw ORCA outputs fundamentally distorts the preference ranking. Because the dynamic programming formulation of ORCA guarantees that the progress score is monotonically non-decreasing over time, $\hat{\reward}_t$ functions as a measure of cumulative progress rather than an instantaneous, per-step reward. Directly summing these cumulative scores across a trajectory segment causes temporal double-counting. This obscures the true advancement achieved within the segment and can lead to inaccurate preference rankings between segments.

To address this issue, we propose converting the progress score into an instantaneous reward signal. Concretely, we define the per-step reward $r_t$ as the \textit{marginal difference} of the ORCA score:
\begin{equation}
r_t = \hat{\reward}_t - \hat{\reward}_{t-1}, \quad  \forall t \in \{2,\dots, H\},
\end{equation}
with the boundary condition $r_1 = \hat{\reward}_1$. This formulation captures the net sequential progress achieved at timestep $t$. Consequently, to evaluate any arbitrary trajectory segment $\sigma = (\state_{t+1}, \dots, \state_{t+L})$, we compute the segment score $R(\sigma)$ by summing these marginal rewards. Crucially, because this forms a telescoping sum, the segment score reduces to the difference between the ORCA score at the final timestep of the segment and the score immediately before the segment begins:
\begin{equation}\label{eq:segment_score}
R(\sigma) = \sum_{k=t+1}^{t+L} r_k = \hat{\reward}_{t+L} - \hat{\reward}_{t}.
\end{equation}
This simplification enables efficient segment evaluation. Subsequently, to derive a preference label for a pair of segments $(\sigma_1, \sigma_2)$, we compare their segment scores and assign the preferred segment $\sigma^w$ and rejected segment $\sigma^l$ as follows:
\begin{equation}\label{eq:comparison}
(\sigma^w, \sigma^l) =
\begin{cases}
(\sigma_1, \sigma_2) & \text{if } R(\sigma_1) > R(\sigma_2) \\
(\sigma_2, \sigma_1) & \text{if } R(\sigma_2) > R(\sigma_1).
\end{cases}
\end{equation}
To ensure a clear learning signal and avoid updating the policy on ambiguous preferences, we filter out pairs where the score difference $|R(\sigma_1) - R(\sigma_2)|$ falls below a discrimination threshold $\epsilon_{\sigma}$.

While step-wise progress can also be estimated using recent progress reward models \cite{zhang2025rewind,tan2025robo,lee2026roboreward,liang2026robometer}, these approaches typically rely on large-scale vision-language architectures (\eg, the 8B-parameter RoboReward \cite{lee2026roboreward}), resulting in substantial computational overhead. Furthermore, their zero-shot evaluations may suffer from hallucinations or poor generalization when the downstream task distribution differs from the pretraining data \cite{tan2025robo}. In contrast, our approach employs a lightweight dynamic programming-based metric that grounds the progress signal directly in domain-specific demonstrations rather than pretrained VLM priors. As a result, \ours{} can generate preference labels significantly more efficiently while ensuring that the resulting supervision remains tightly aligned with the underlying task dynamics.

\begin{algorithm}[t]
\small
\caption{Pseudo-code for \fullours{} (\ours{})}
\label{alg:pseudo_code}
\begin{algorithmic}[1]
    \STATE \textbf{Input}: Dataset of demonstrations $\expertdataset$, dataset of rollouts $\rolloutdataset$, number of expert references $M$, number of queries $N$, discrimination threshold $\epsilon_{\sigma}$, pretrained policy $\pi_{\theta}$.
    \STATE \textbf{Initialize}: Select reference set $\{\tilde{\tau}_m\}_{m=1}^M \subset \expertdataset$, $\mathcal{D} = \expertdataset \cup \rolloutdataset$, preference dataset $\prefdataset \leftarrow \emptyset$.
    \STATE Construct progress-labeled dataset $\hat{\mathcal{D}}$ by calculating ORCA score for trajectories in $\mathcal{D}$ using Eq.~(\ref{eq:orca_reward}).
    \FOR{$i = 1$ \TO $N$}
        \STATE Sample trajectory segments $(\sigma_1, \sigma_2) \sim \hat{\mathcal{D}}$.
        \STATE Compute segment scores $R(\sigma_1)$ and $R(\sigma_2)$ using Eq.~(\ref{eq:segment_score}) and Eq.~(\ref{eq:aggregate_score}).
        \IF{$|R(\sigma_1) - R(\sigma_2)| > \epsilon_{\sigma}$}
            \STATE Assign preferred segment $\sigma^w$ and rejected segment $\sigma^l$ based on Eq.~(\ref{eq:comparison}).
            \STATE $\prefdataset \leftarrow \prefdataset \cup \{(\sigma^w, \sigma^l)\}$
        \ENDIF
    \ENDFOR
    \STATE Fine-tune policy $\pi_{\theta}$ with dataset $\prefdataset$ utilizing the CPL objective in Eq.~(\ref{eq:cpl_loss}).
\end{algorithmic}
\end{algorithm}
\setlength{\textfloatsep}{0.15in}

\paragraph{Multi-Demonstration Aggregation.} In robotic control tasks, there are often multiple valid strategies for achieving the same goal. Relying on a single demonstration to generate preference labels may inadvertently penalize a rollout that successfully completes the task using an alternative yet equally valid behavioral mode. To capture this multimodality, we evaluate the segment score using a subset of $M$ demonstrations. Specifically, we compute the segment score $R(\sigma, \tilde{\tau}_m)$ independently for each demonstration $\tilde{\tau}_m$, and then extract the final score $R(\sigma)$ via max-pooling across this subset:
\begin{equation}\label{eq:aggregate_score}
R(\sigma) = \max_{m \in \{1\dots M\}} R(\sigma, \tilde{\tau}_m).
\end{equation}
Once the aggregated scores for two segments $\sigma_1$ and $\sigma_2$ are computed, we apply the discrimination threshold $\epsilon_{\sigma}$ and the comparison in Eq.~(\ref{eq:comparison}) to construct the final preference pair $(\sigma^w, \sigma^l)$. Crucially, our approach applies this max-pooling operation at the segment level, rather than over the sum of the entire trajectory as is commonly done in prior sequence-matching methods \cite{fu2024robot,huey2025imitation}. This segment-wise aggregation provides a key behavioral advantage by enabling the framework to properly reward composite trajectories. For instance, a learner may follow one expert's strategy during an initial segment and transition seamlessly to an alternative strategy in a later segment, receiving reliable progress signals without being forced to merely imitate a single global demonstration. 

\subsection{Implementation Details}
\vspace{-0.05in}
Since ORCA evaluates high-dimensional visual observations within trajectories, we first project raw image frames into a semantic latent space using a pretrained visual encoder before computing pairwise distances. Specifically, we employ R3M \cite{nair2022r3m} as the visual backbone, which in our preliminary experiments provided more robust semantic alignment than the ResNet50 \cite{he2016deep} used in the original ORCA implementation \cite{huey2025imitation}. By applying \ours{} to score these trajectory segments, we construct a preference dataset $\prefdataset$. This dataset is subsequently utilized to fine-tune the pre-trained policy via the CPL objective defined in Eq.~(\ref{eq:cpl_loss}). The overall procedure is summarized in Algorithm \ref{alg:pseudo_code}.

\vspace{-0.1in}
\section{Experiments}
\label{sec:experiments}
\vspace{-0.1in}
In this section, we conduct comprehensive experiments across a suite of complex manipulation tasks to investigate the following questions: (1) Can \ours{} synthesize reliable preference labels to guide policy learning? (2) How effectively does our framework improve the performance of pretrained generative policies compared to supervised fine-tuning (SFT) baselines? (3) How do critical hyperparameters influence the overall performance of \ours{}?

\subsection{Experimental Setups}
\label{subsec:setup}
\vspace{-0.05in}
To evaluate the effectiveness of our method, we adopt the Robomimic benchmark \cite{mandlekar2021matters}, which consists of five tabletop manipulation tasks: \emph{Lift}, \emph{Can}, \emph{Square}, \emph{Transport}, and \emph{Tool Hang}. These tasks cover diverse manipulation skills and long-horizon behaviors, ranging from simple pick-and-place tasks to precise insertion tasks. To evaluate the generality of our framework, we conduct experiments using both Diffusion Policy \cite{chi2023diffusion} and Rectified Flow policies \cite{zhang2025reinflow}. Following prior work \cite{miao2025uncertainty}, for each task, the reference policy $\pi_{\text{ref}}$ is trained to achieve approximately 40\% success rate, after which we collect 500 rollout trajectories to construct the rollout dataset. We further retain only 40\% of the original demonstration dataset, resulting in approximately 120 demonstrations per task (80 for \textit{Tool Hang}). We use the \textit{mh} datasets from Robomimic for all tasks except \textit{Tool Hang}, which only provides a \textit{ph} dataset. All methods are trained using low-dimensional state observations. Visual observations are used solely by PGPG for preference generation. We evaluate the performance of the fine-tuned policy after 1000 training epochs. Results are averaged over 50 episodes across three random seeds, as in \cite{liu2025bidirectional}. We compare \ours{} against the following baselines:

\begin{itemize}[leftmargin=1em,topsep=0pt,itemsep=0pt]  
    \item \textbf{Base}: The pretrained generative policy $\pi_{\text{ref}}$ (diffusion or flow-matching) prior to any fine-tuning.
    
    \item \textbf{Mixed BC}: Standard behavioral cloning that fine-tunes the base policy using the combined dataset of expert demonstrations and all uncurated rollouts.
    
    \item \textbf{Filtered BC} \cite{mirchandani2024so,chen2025curating}: Behavioral cloning fine-tuned using expert demonstrations and only successful rollout trajectories. Crucially, these successful rollouts are filtered using the benchmark's oracle success criteria, which represents privileged environmental information inaccessible to our method.
    
    \item \textbf{$\text{CPL}_{KL}$} \cite{hejna2024contrastive}: A contrastive preference learning baseline that follows our exact training setup, but derives preference labels directly from the ground-truth benchmark reward functions \cite{lee2021b} rather than our progress metric. For environments with dense rewards (Lift, Can, and Square), we use 1000 preference labels; for the remaining sparse-reward tasks, we use 5000 labels.
    
    \item \textbf{DPPO} \cite{ren2025diffusion}: A reinforcement learning baseline that fine-tunes the base diffusion policy using a PPO-based variant \cite{ren2025diffusion}. For a fair comparison, the pretrained model in this baseline achieves comparable performance to our reference policy, and we restrict it to approximately the same number of additional rollout trajectories used by \ours{}.
\end{itemize}

\begin{figure*}[t]
\begin{center}
    \includegraphics[width=0.6\textwidth]{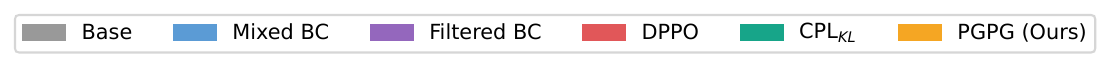}
    \begin{subfigure}[b]{\textwidth}
    \includegraphics[width=\textwidth]{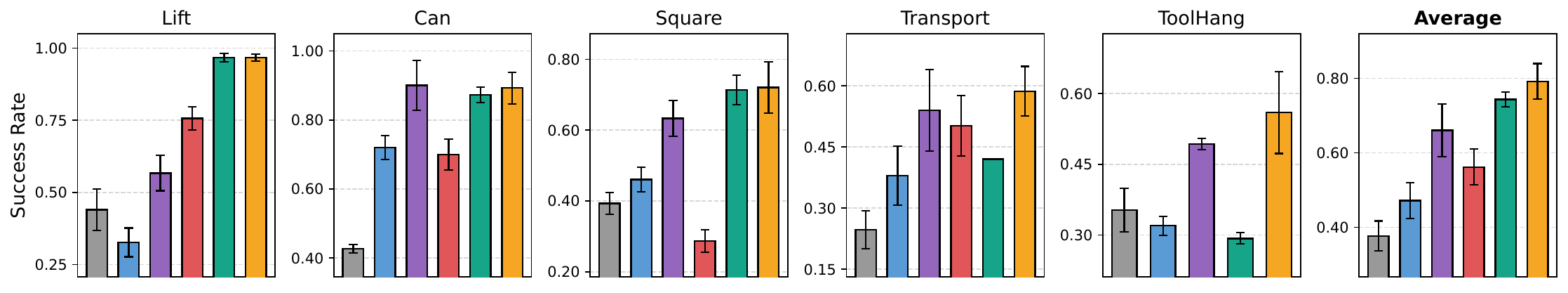}
    \caption{Diffusion policy}
    \label{fig:dp_result}
    \end{subfigure}
    \begin{subfigure}[b]{\textwidth}
    \includegraphics[width=\textwidth]{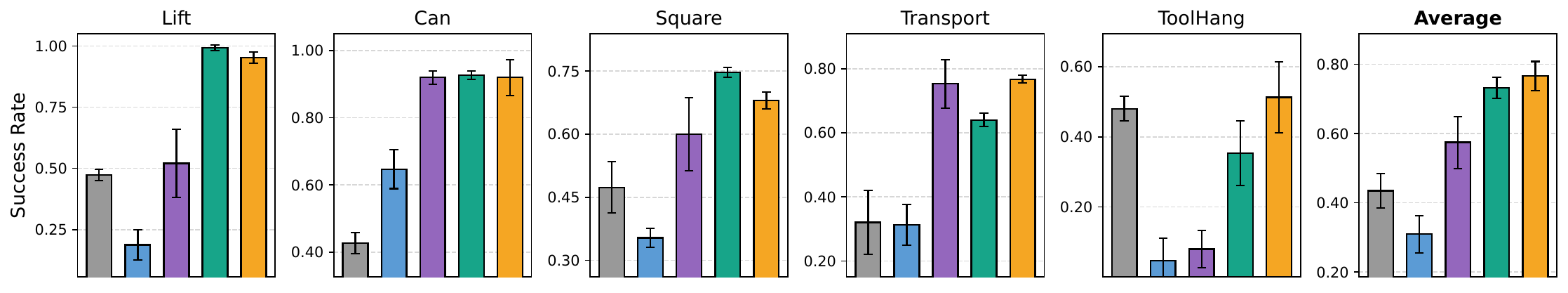}
    \caption{Flow-matching policy}
    \label{fig:fp_result}
    \end{subfigure}
\end{center}
\vspace{-0.05in}
\caption{Success rates of methods on five Robomimic manipulation tasks. Error bars denote standard deviation across seeds. DPPO is applied only to the diffusion policy.}
\label{fig:main_results}
\end{figure*}

\subsection{Main Results}

As shown in Figure~\ref{fig:main_results}, \ours{} achieves the highest average success rate across all tasks, significantly outperforming non-privileged baselines such as Mixed BC and DPPO. In fact, Mixed BC yields only marginal improvements and occasionally degrades base policy performance. Among the baselines that use privileged information, Filtered BC and $\text{CPL}_{KL}$, the preference-based fine-tuning of $\text{CPL}_{KL}$ achieves stronger results. This indicates that selectively ranking trajectory quality provides a more effective learning signal than directly mixing successful rollouts with expert demonstrations.

Notably, on tasks with dense ground-truth rewards, \ours{} matches the performance of the reward-guided $\text{CPL}_{KL}$ baseline (with the sole exception of the Flow Policy on \textit{Square}). This confirms that our automatically generated preferences effectively replace manually designed reward functions. Furthermore, \ours{} consistently outperforms $\text{CPL}_{KL}$ on sparse-reward tasks (\textit{Transport} and \textit{ToolHang}). We attribute this advantage to the limitations of sparse rewards, which assign nearly identical supervision to distinct segment pairs, failing to capture subtle differences in progress. In contrast, our progress-guided preferences deliver dense, temporally structured supervision, leading to superior policy fine-tuning.

\subsection{Ablation Studies}
Unless otherwise stated, all ablation studies are conducted with the diffusion policy.

\paragraph{Choice of Alignment Metric.}
We investigate the effectiveness of different sequence alignment metrics for preference generation by applying our Marginal Progress Scoring (MPS) to Temporal OT \cite{fu2024robot}, DTW \cite{sakoe1978dynamic}, and the ORCA baseline \cite{huey2025imitation}. As shown in Table~\ref{tab:sequence_matching}, MPS yields large, consistent gains only when combined with ORCA. When applied to other metrics, MPS produces inconsistent results: it generally improves Temporal OT (except on Square) but degrades DTW on three of five tasks. This suggests that the gains arise from the interaction between marginal scoring and ORCA's ordered-coverage structure, rather than from MPS or ORCA alone. This finding is consistent with ORCA's analysis \cite{huey2025imitation}: DTW and Temporal OT do not reliably capture full subgoal coverage, resulting in an unreliable ranking signal. In contrast, ORCA's score is provably monotone in the number of subgoals covered.

\begin{table}[t]
\centering
\caption{Comparison of fine-tuning performance using preference labels generated by different sequence-matching metrics within our framework. Best results and those within one standard deviation are bolded.}
\begin{tabular}{lccccc}
\toprule
Method & Lift & Can & Square & Transp. & ToolHang \\
\midrule
Temporal OT      & 20.7\stdv{1.2}  & 58.7\stdv{5.8} & 62.7\stdv{3.1}  & 36.7\stdv{6.1} & 32.0\stdv{5.3} \\
Temporal OT + MPS    & 82.0\stdv{2.0}  & 72.0\stdv{5.3} & 51.3\stdv{10.1} & 43.3\stdv{5.0} & 41.3\stdv{8.3} \\
DTW      & 90.7\stdv{3.1}  & 70.0\stdv{5.3} & \textbf{72.7}\stdv{1.2}  & 45.3\stdv{11.0} & 30.0\stdv{8.7} \\
DTW + MPS    & 80.0\stdv{5.3}  & 78.0\stdv{7.2} & 62.7\stdv{7.0}  & 38.7\stdv{2.3} & 31.3\stdv{7.0} \\
ORCA     & 32.0\stdv{2.0}  & 70.7\stdv{5.0} & 42.0\stdv{3.5}  & 47.3\stdv{4.6} & 41.3\stdv{3.1} \\
ORCA + MPS (Ours) & \textbf{96.7}\stdv{1.2} & \textbf{89.3}\stdv{4.6} & \textbf{72.0}\stdv{7.2} & \textbf{58.7}\stdv{6.1} & \textbf{56.0}\stdv{8.7} \\
\bottomrule
\end{tabular}
\label{tab:sequence_matching}
\end{table}

\begin{figure*}[t]
\begin{subfigure}[t]{0.32\textwidth}
    \includegraphics[width=\textwidth]{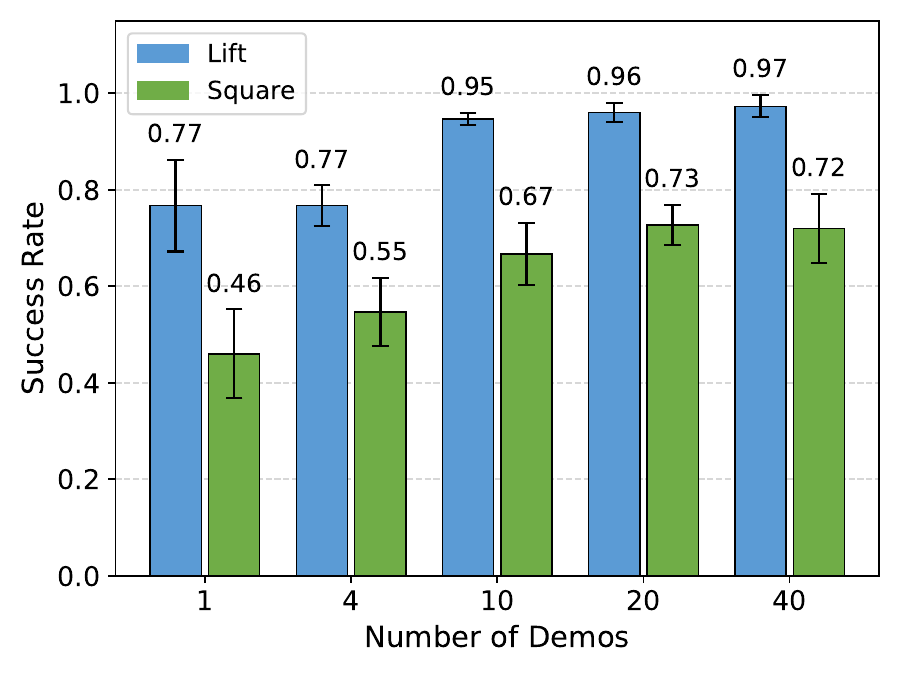}
    \caption{Varying Demonstrations}
    \label{fig:n_demo_impact}
\end{subfigure}
\hfill
\begin{subfigure}[t]{0.32\textwidth}
    \includegraphics[width=\textwidth]{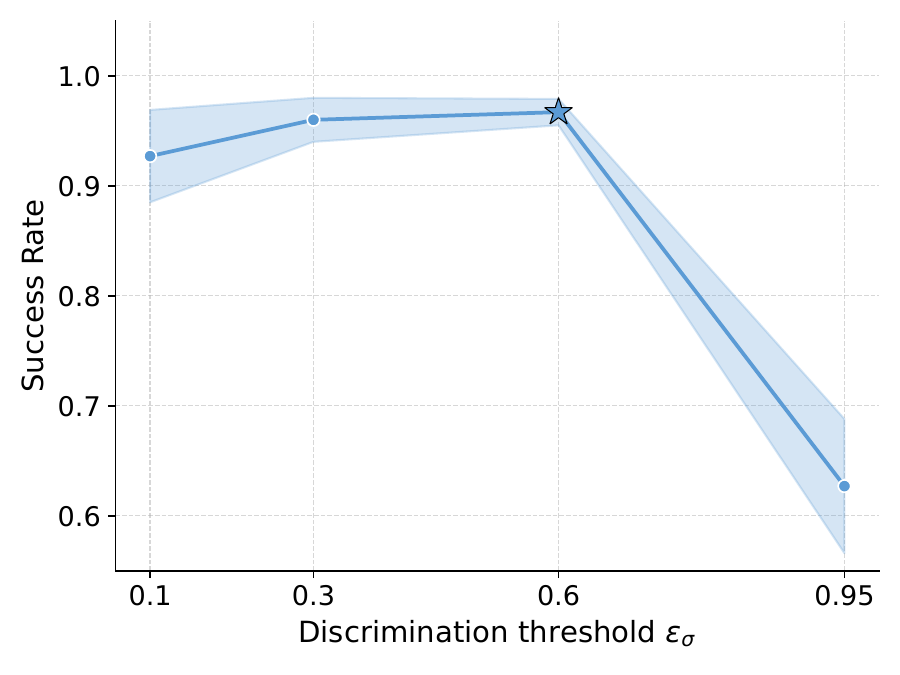}
    \caption{Effect of $\epsilon_{\sigma}$}
    \label{fig:segment_margin}
\end{subfigure}
\hfill
\begin{subfigure}[t]{0.32\textwidth}
    \includegraphics[width=\textwidth]{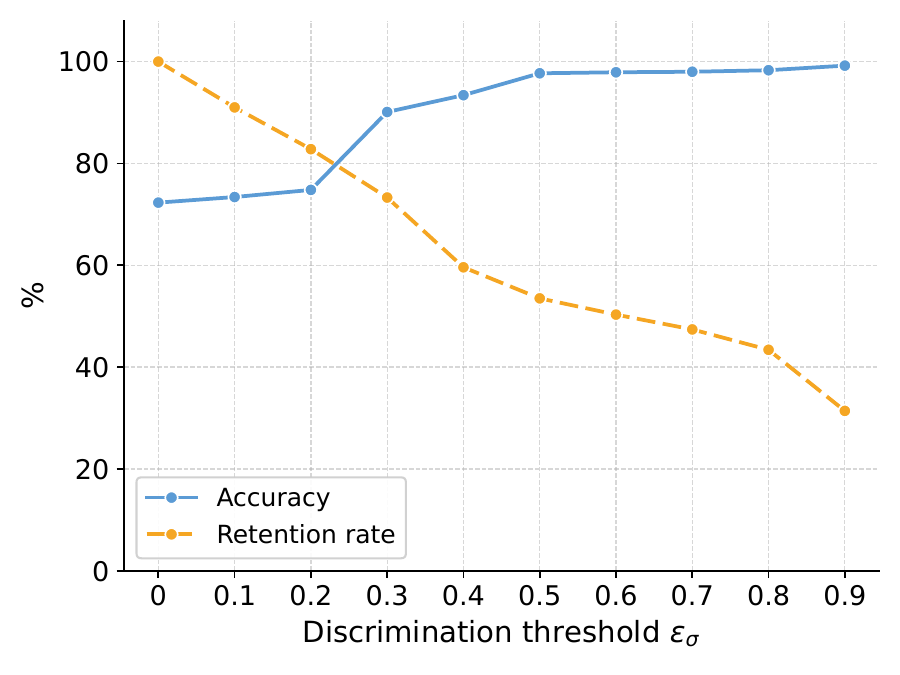}
    \caption{Accuracy and Retention Rate}
    \label{fig:tradeoff_lift}
\end{subfigure}
\caption{Ablation study across different design choices.}
\vskip -0.1in
\label{fig:ablation}
\end{figure*}

\paragraph{Impact of the Number of Demonstrations.} We evaluate the data efficiency of \ours{} by varying the number of expert demonstrations used for label synthesis (Figure \ref{fig:n_demo_impact}). The results indicate rapid performance improvements as the number of demonstrations increases from 1 to 10 across both the \emph{Lift} and \emph{Square} tasks. Notably, the success rate saturates after 10--20 demonstrations, highlighting that our framework can achieve near-optimal policy alignment without the burden of large-scale expert data collection.

\vspace{-0.05in}
\paragraph{Quality of Pseudo-Preference Labels.} We evaluate the accuracy and retention rate of our synthesized preference labels against ground-truth preferences under varying discrimination thresholds ($\epsilon_{\sigma}$). As illustrated in Figures~\ref{fig:segment_margin} and \ref{fig:tradeoff_lift}, adjusting this threshold reveals a critical trade-off between label reliability and dataset scale. While larger thresholds improve label accuracy by filtering out ambiguous segment pairs, they simultaneously reduce the amount of usable training data. Consequently, downstream policy performance peaks at a moderate threshold ($\epsilon_{\sigma}=0.6$), which effectively balances label quality and data retention. In contrast, excessively large thresholds ($\epsilon_{\sigma}=0.95$) severely reduce the available training data, leading to a sharp performance degradation despite near-perfect label accuracy. Further ablations are provided in the Appendix.

\vspace{-0.05in}
\section{Conclusion}
\vspace{-0.1in}
We presented \fullours{} (\ours{}), an unsupervised framework that aligns generative control policies without relying on human annotations or explicit reward engineering. By leveraging ordered coverage alignment, our method autonomously evaluates sequential progress against a small set of expert demonstrations, translating suboptimal rollouts into robust pseudo-preference labels. Empowered by key enhancements to ensure label reliability, \ours{} significantly improves the performance of pretrained generative policies on challenging, long-horizon manipulation tasks, outperforming standard SFT baselines.
\vspace{-0.05in}
\paragraph{Limitations.}
First, \ours{} relies heavily on the robustness of the pre-trained visual encoder; future work will explore online fine-tuning to support cross-embodiment alignment where temporal mismatches are severe. Second, visual sequence matching remains susceptible to task misspecification under heavy occlusion. We plan to mitigate this by integrating richer modalities, such as depth or tactile sensing. Finally, our Marginal Progress Scoring assumes a largely monotonic progression toward the goal, which may struggle to accurately score long-horizon tasks that require strategic backtracking or non-quasistatic behavior.

\section*{Acknowledgements}
This work was supported by Institute of Information \& communications Technology Planning \& Evaluation (IITP) grant funded by the Korea government (MSIT) (No.RS-2021-II211381, Development of Causal AI through Video Understanding and Reinforcement Learning, and Its Applications to Real Environments) and the National Research Foundation of Korea (NRF) grant funded by the Korea government(MSIT) (RS-2025-24742969, Intelligent Robotic System using Continual Learning and Multimodal Language Model based Multi Attribute Feedback). Duy M. H. Nguyen and Daniel Sonntag also acknowledge the support of the Federal Ministry of Research, Technology, and Space (BMFTR) under grant number 16IW26002 (AMENABLE), the Lower Saxony Ministry of Science and Culture (MWK) in the Zukunft.niedersachsen program, and the Endowed Chair of AAI at the University of Oldenburg.

\bibliography{references}  

\clearpage
\appendix

\section{Appendix}

\etocsettocstyle{}{} 
\localtableofcontents 
\vspace{2em} 

\subsection{Additional Experiments and Results}

\subsubsection{Learning from Different Numbers of Rollouts}

We investigate how the size of the autonomous rollout dataset impacts the fine-tuning performance of \ours{}. As shown in Table \ref{tab:ablation_rollout}, decreasing the number of uncurated rollouts leads to a slight drop in policy performance across both tasks. For the \emph{Square} task, the success rate declines steadily as the dataset shrinks, dropping from a peak of $72.0\%$ at 500 rollouts down to $65.7\%$ at 125 rollouts. Conversely, the \emph{Lift} task demonstrates higher resilience to reduced data; it maintains near-optimal alignment ($96.0\%$) with just 250 rollouts, and only experiences a slight decrease to $92.3\%$ when the dataset is further reduced to 125 rollouts. This trend demonstrates that \ours{} effectively extracts dense preference gradients from suboptimal data, achieving robust behavioral alignment without the need for large-scale data collection.

\begin{table}[h]
\centering
\vskip -0.1in
\caption{\ours{}'s performance using different numbers of rollouts (\ie, $|\rolloutdataset|$).}
\begin{tabular}{lcc}
\toprule
Number of rollouts & Lift & Square \\
\midrule
500     & 96.7 \stdv{1.2} & 72.0 \stdv{7.2}   \\
375     & 94.7 \stdv{3.2} & 71.7 \stdv{2.5}   \\
250     & 96.0 \stdv{3.0} & 69.7 \stdv{6.1}  \\
125     & 92.3 \stdv{4.5} & 65.7 \stdv{3.1}  \\
\bottomrule
\end{tabular}
\label{tab:ablation_rollout}
\vskip -0.1in
\end{table}

\subsubsection{Learning from Different Numbers of Demonstrations}

In the main experiments, we found that 120 demonstrations were sufficient to achieve strong performance for both the baselines and our method. We further evaluate smaller demonstration budgets in Table~\ref{tab:demos}, while keeping the preference-generation procedure unchanged (\ie, using a fixed number of reference demonstrations, as described in Section \ref{sec:detail_exp_setting}). With only 40 demonstrations, PGPG's performance remains nearly unchanged on \emph{Lift} and \emph{Can}, with a moderate drop on \emph{Square}. Even with \emph{no} demonstrations (pure rollouts), PGPG still improves over the base policy on \emph{Can} and \emph{Square}.

\begin{table}[h]
\centering
\vskip -0.15in
\caption{PGPG's performance using different numbers of demonstrations (\ie, $|\expertdataset|$).}
\begin{tabular}{lccc}
\toprule
Number of demos & Lift & Can & Square \\
\midrule
Base policy         & 44.0\stdv{7.2} & 42.7\stdv{1.2} & 39.3\stdv{3.1} \\
0 (pure rollouts)   & 32.7\stdv{4.6} & 64.7\stdv{6.4} & 51.3\stdv{5.0} \\
40                  & 95.3\stdv{1.2} & 88.0\stdv{4.0} & 56.7\stdv{6.1} \\
120                 & 96.7\stdv{1.2} & 89.3\stdv{4.6} & 72.0\stdv{7.2} \\
\bottomrule
\end{tabular}
\label{tab:demos}
\vskip -0.1in
\end{table}

\subsubsection{Impact of Visual Encoder}

We vary the visual encoder within the PGPG framework. The results are reported in Table~\ref{tab:encoder}. The R3M encoder used in our main experiments performs best, while VC-1 \cite{majumdar2023we} is competitive and the general vision-language SigLIP2 \cite{tschannen2025siglip} is weaker. Consistent with ORCA \cite{huey2025imitation}, the choice of encoder is a practical, task-dependent consideration. 

\begin{table}[h]
\centering
\vskip -0.15in
\caption{PGPG performance with different visual encoders.}
\begin{tabular}{lccc}
\toprule
Encoder & Lift & Can & Square \\
\midrule
Base policy        & 44.0\stdv{7.2} & 42.7\stdv{1.2} & 39.3\stdv{3.1} \\
PGPG w/ VC-1       & 96.0\stdv{4.0} & 83.3\stdv{4.6} & 52.7\stdv{3.1} \\
PGPG w/ SigLIP2    & 68.7\stdv{7.0} & 58.0\stdv{8.7} & 42.7\stdv{7.0} \\
PGPG w/ R3M (Ours) & 96.7\stdv{1.2} & 89.3\stdv{4.6} & 72.0\stdv{7.2} \\
\bottomrule
\end{tabular}
\label{tab:encoder}
\vskip -0.1in
\end{table}

\subsubsection{Impact of Segment Length}

Table~\ref{tab:ablation_segsize} illustrates the impact of segment length on the \emph{Square} task. Performance is notably lower at $L=100$, suggesting that segments must be sufficiently long to capture meaningful physical transitions rather than trivial movements. However, at $L = 150$, performance increases substantially and plateaus through $L=200$. This indicates that once the segment length is adequate to encapsulate meaningful sub-behaviors, preference learning with \ours{} remains highly stable.

\begin{table}[h]
\centering
\caption{Performance of \ours{} with different segment lengths $L$ on Square task.}
\begin{tabular}{lcccc}
\toprule
$L$             & 100 & 150 & 200  \\
\midrule
Success Rate    &  54.3 \stdv{7.6}  & 71.3 \stdv{7.1}  &  72.0 \stdv{7.2}  \\
\bottomrule
\end{tabular}
\label{tab:ablation_segsize}
\vskip -0.1in
\end{table}

\subsubsection{Multi-Demonstration Aggregation Operator}

As shown in Table \ref{tab:aggregation_operator}, the $\max$ operator consistently outperforms $\text{mean}$ pooling across both tasks. This validates our hypothesis: averaging progress scores across diverse demonstrations heavily penalizes valid, multimodal behaviors by forcing segments to match a potentially non-existent ``blurry'' trajectory. Conversely, the $\max$ operator explicitly preserves multimodality by rewarding a segment if it aligns with any single valid expert mode, effectively mitigating spurious matches and inconsistent preference rankings.

\begin{table}[h]
\centering
\vskip -0.1in
\caption{Performance of \ours{} with different aggregation operators across demonstrations.}
\begin{tabular}{lcc}
\toprule
Aggregation             & Lift & Square \\
\midrule
Mean    &  90.7 \stdv{2.3}  &  52.7 \stdv{2.1} \\
Max     &  96.7 \stdv{1.2}  &  72.0 \stdv{7.2} \\
\bottomrule
\end{tabular}
\label{tab:aggregation_operator}
\vskip -0.1in
\end{table}

\subsubsection{Comparison with Vision-Language and Progress Reward Baselines}
To compare PGPG with other preference-generation methods that do not require querying large VLMs or LLMs, we evaluate three approaches within our exact pipeline, replacing only the label source: frame-based CLIP \cite{rocamonde2024vision}, video-based S3D \cite{sontakke2023roboclip}, and the Robometer progress reward model \cite{liang2026robometer}. As shown in Table~\ref{tab:vlm_reward}, PGPG performs on par with or better than all baselines on all considered tasks. Robometer performs competitively only on \emph{Can} and struggles on the precision and long-horizon tasks, while being approximately $31\times$ slower than \ours{} (see Section \ref{sec:pgpg_robometer_compute}). CLIP and S3D measure similarity between visual observations and the task description but do not explicitly account for ordered subgoal completion, a key property captured by ORCA.

\begin{table}[h]
\centering
\vskip -0.15in
\caption{Comparison with Vision-Language and Progress Reward Baselines. Best results and those within one standard deviation are bolded.}
\begin{tabular}{lccccc}
\toprule
Method & Lift & Can & Square & Transp. & ToolHang \\
\midrule
CLIP          & 36.0\stdv{7.2} & 62.0\stdv{6.0} & 58.0\stdv{8.7} & 21.3\stdv{4.2} & 30.7\stdv{9.0} \\
S3D           & 53.3\stdv{3.1} & 60.0\stdv{3.5} & 62.8\stdv{4.6} & 26.7\stdv{4.2} & \textbf{54.0}\stdv{8.7} \\
Robometer     & 38.7\stdv{8.1} & \textbf{85.3}\stdv{3.1} & 40.0\stdv{3.5} & 28.0\stdv{6.9} & 31.3\stdv{6.1} \\
PGPG (Ours) & \textbf{96.7}\stdv{1.2} & \textbf{89.3}\stdv{4.6} & \textbf{72.0}\stdv{7.2} & \textbf{58.7}\stdv{6.1} & \textbf{56.0}\stdv{8.7} \\
\bottomrule
\end{tabular}
\label{tab:vlm_reward}
\end{table}

\subsubsection{Computational Efficiency}\label{sec:pgpg_robometer_compute}
\begin{figure*}[h]
\begin{center}
    \includegraphics[width=0.4\linewidth]{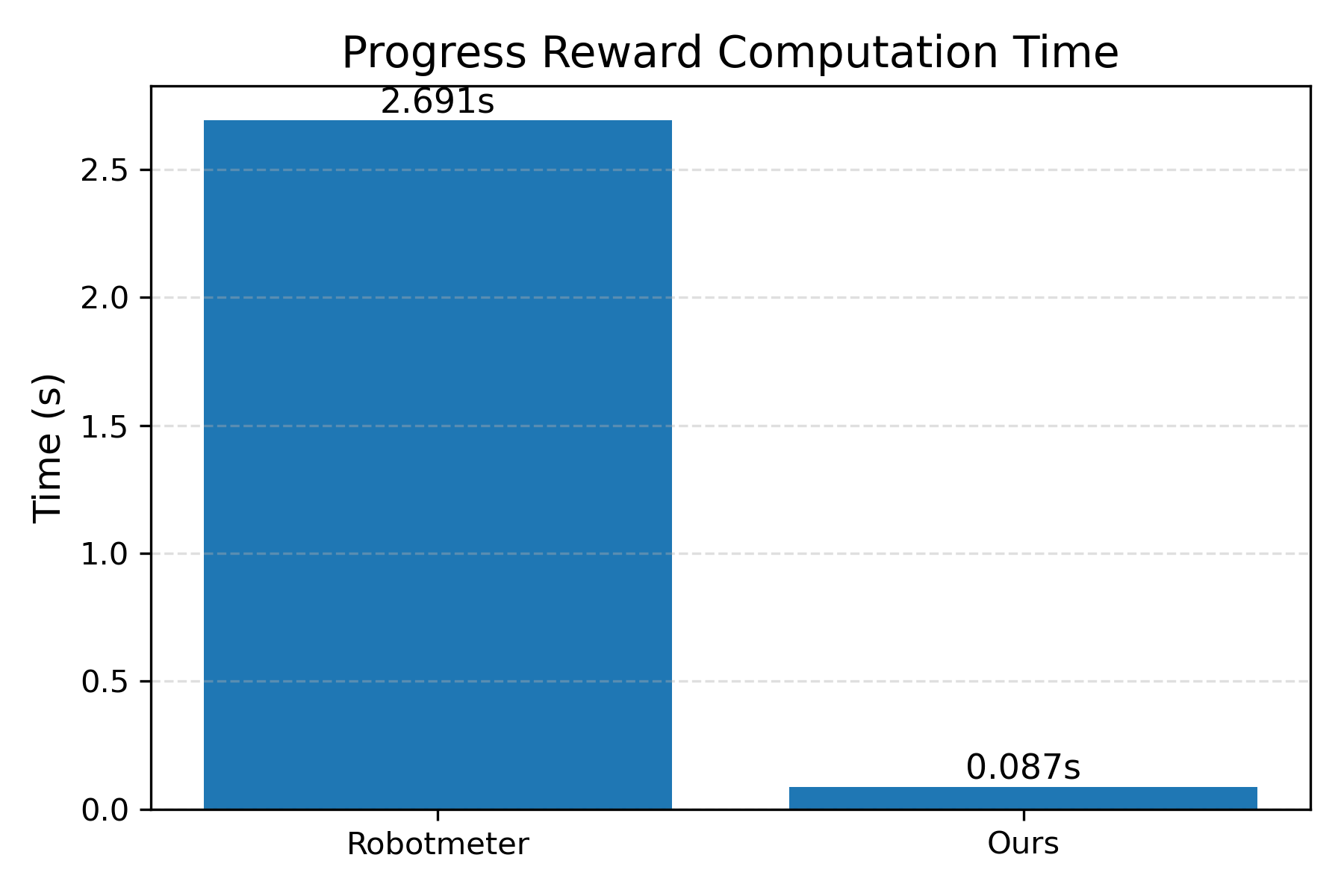}
\end{center}
\vskip -0.1in
\caption{Average wall-clock time required to compute progress rewards per trajectory, evaluated over 50 rollouts. By leveraging the ORCA progress metric, \ours{} is nearly $31\times$ faster than the Robometer \cite{liang2026robometer} baseline, successfully eliminating the high inference latency typically associated with progress reward models.}
\vskip -0.1in
\label{fig:prm_comparison}
\end{figure*}

Figure \ref{fig:prm_comparison} compares the inference latency of \ours{} against Robometer \cite{liang2026robometer}. While PRMs and VLMs introduce significant wall-clock latency per rollout due to heavy network inference, leveraging the ORCA metric \cite{huey2025imitation} allows \ours{} to evaluate trajectory progress significantly faster.

\subsubsection{Applying \ours{} with GR00T N1.7}
The experiments in the main paper fine-tune task-specific diffusion and flow-based policies, each trained on a single Robomimic task. To test whether \ours{} generalizes beyond this setting, we apply it to GR00T N1.7~\cite{bjorck2025gr00t}, a large pretrained vision-language-action (VLA) model, on the multi-task LIBERO-10 benchmark~\cite{liu2023libero}. This evaluation further demonstrates the effectiveness of our method in three aspects:
\begin{itemize}
  \item[(i)] \textbf{Multi-task learning.} A single language-conditioned policy is fine-tuned on all ten tasks at once. Preferences are generated for each task against that task's own demonstrations and then pooled into one fine-tuning run, without task-specific models or hyperparameters.
  \item[(ii)] \textbf{Pretrained VLA.} The policy is a 3B-parameter model in which a vision-language backbone conditions a flow-matching action head. \ours{} needs only the flow-matching loss of the action head, which takes the place of the log-likelihood in $\text{CPL}_{KL}$, so it applies to the VLA without architectural changes.
  \item[(iii)] \textbf{Few-shot data.} Only 5 demonstrations and 20 policy rollouts are available per task, compared with about 120 demonstrations and 500 rollouts per task in the main experiments.
\end{itemize}

\begin{figure}[h]
  \centering
  \includegraphics[width=\linewidth]{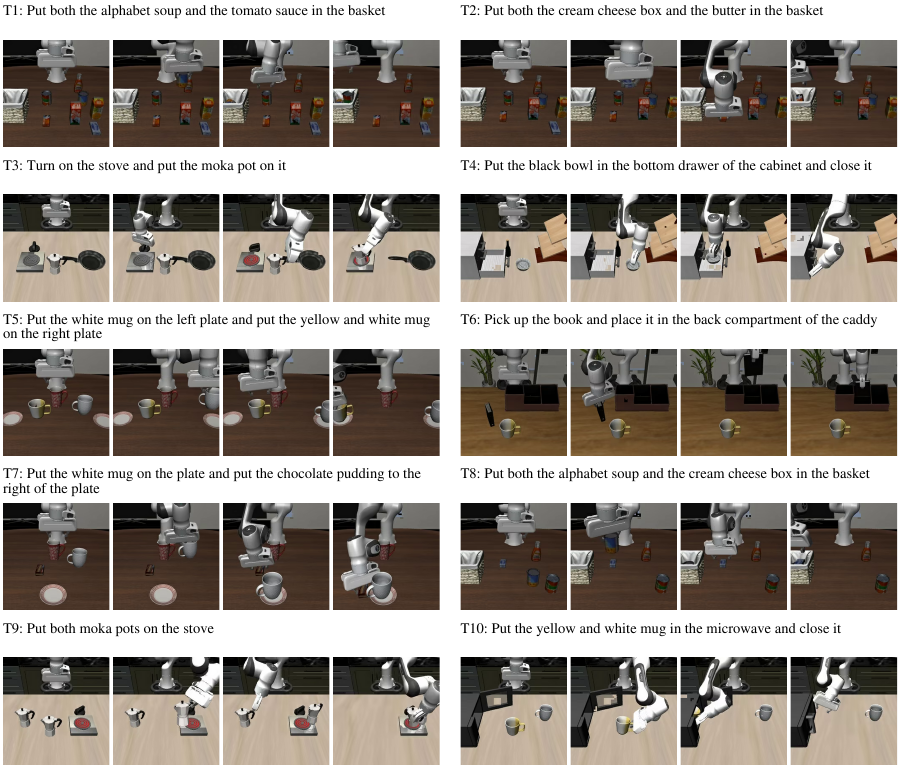}
  \caption{LIBERO-10 tasks: Each row of frames shows one of the five available expert demonstrations for a task, progressing from the initial state (left) to task completion (right).}
  \label{fig:libero10_demos}
\end{figure}

\paragraph{Benchmark and Dataset.}
LIBERO-10 (also known as LIBERO-Long) consists of ten long-horizon, language-conditioned manipulation tasks for a simulated Franka Panda arm, each chaining two sub-goals (illustrated in Figure~\ref{fig:libero10_demos}).
The policy takes as input a third-person and a wrist-camera image ($256\times256$), the end-effector state, and the instruction.
We study a few-shot learning setting in which only five teleoperated demonstrations are available per task (50 in total, from the LIBERO-10 dataset with no-op actions removed). The pretrained policy collects 20 rollouts per task (200 in total), of which 75 are successful.

\paragraph{Pretrained Policy.}
GR00T N1.7 combines a vision-language model (VLM) backbone (Cosmos-Reason2-2B) with a Diffusion Transformer action head trained with flow matching, which predicts action chunks of length $T_p=16$. We obtain the pretrained policy $\pi_{\mathrm{ref}}$ by fine-tuning the released GR00T N1.7 checkpoint on the 50 demonstrations for 5000 iterations, using a batch size of 16 and a learning rate of $10^{-4}$, while keeping the VLM backbone frozen.

\paragraph{Implementation of \ours{}.}
Pseudo-preference labels are generated as in the main paper: ORCA progress \cite{huey2025imitation} is computed on R3M features \cite{nair2022r3m} of the third-person image by comparing it with the five demonstrations of the same task, and each segment is scored by its marginal progress, maximized over the demonstrations.
Segments span 160 environment steps (ten action chunks). For each task we sample 5000 demonstration-vs-rollout queries and 5000 rollout-vs-rollout queries, require both segments in each query to start at the same progress level ($\pm0.05$), and discard queries whose score gap is below $0.1$, leaving 47878 labeled pairs.
The policy is fine-tuned with $\text{CPL}_{KL}$ \cite{hejna2024contrastive} using $\pi_{\mathrm{ref}}$ as the reference. Because the action head is a flow-matching model, the per-chunk log-likelihood ratio is replaced by the difference between the velocity-prediction errors of $\pi_\theta$ and $\pi_{\mathrm{ref}}$, computed with shared noise and flow time ($\alpha=0.015$, discount $0.999$, weight $0.25$ on the losing segment, with per-step terms clipped at $0.2$).
Only the action head is updated using cached backbone features. We train for 8000 steps with AdamW (learning rate $10^{-4}$, cosine schedule with 5\% warm-up) and 48 preference pairs per batch on four NVIDIA A100 80 GB GPUs.

\paragraph{Baseline and Evaluation.}
Filtered BC fine-tunes $\pi_{\mathrm{ref}}$ using the flow-matching objective on the demonstrations and 75 successful rollouts, with the same learning rate, batch size, and number of training steps. Unlike \ours{}, it relies on the simulator's success signal to select rollouts.
All policies are evaluated at their final checkpoints over 50 episodes per task for each of three seeds. For a given seed, every method starts from the same initial states. The policies execute the first eight actions of each predicted chunk, and an episode ends upon success or after 720 steps.

\paragraph{Results.} The results are shown in Table~\ref{tab:pgpg_on_vla}. \ours{} improves the average success rate from 50.5\% for the pretrained policy to 72.1\%, compared with 64.8\% for Filtered BC. This corresponds to gains of $21.5\pm1.8$ percentage points over the pretrained policy and $7.3\pm3.6$ percentage points over Filtered BC. \ours{} achieves the best performance on seven of the ten tasks, with the largest gains on T1 ($18.0\%\rightarrow64.7\%$), T8 ($40.0\%\rightarrow84.0\%$) and T2 ($58.0\%\rightarrow94.7\%$). Filtered BC performs better on T5 and T6. Neither method solves T9 (success rates of at most 3.3\%), where none of the 20 rollouts collected by the pretrained policy succeeds.

\begin{table}[h]
  \centering
  \caption{Success rate (\%) of methods on LIBERO-10. Mean $\pm$ standard deviation over three seeds with 50 episodes per task each. The best result for each task is shown in bold.}
  \label{tab:pgpg_on_vla}
  \footnotesize
  \setlength{\tabcolsep}{4pt}
  \begin{tabularx}{\linewidth}{@{}l>{\raggedright\arraybackslash}Xccc@{}}
    \toprule
    Task &  Instruction & Pretrained & Filtered BC & \ours{} \\
    \midrule
    T1  & put both the alphabet soup and the tomato sauce in the basket                          & 18.0\,{\scriptsize$\pm$\,5.3}  & 37.3\,{\scriptsize$\pm$\,3.1}  & \textbf{64.7}\,{\scriptsize$\pm$\,7.6} \\
    T2  & put both the cream cheese box and the butter in the basket                             & 58.0\,{\scriptsize$\pm$\,2.0}  & 76.7\,{\scriptsize$\pm$\,1.2}  & \textbf{94.7}\,{\scriptsize$\pm$\,3.1} \\
    T3  & turn on the stove and put the moka pot on it                                           & 80.0\,{\scriptsize$\pm$\,3.5}  & 76.7\,{\scriptsize$\pm$\,1.2}  & \textbf{87.3}\,{\scriptsize$\pm$\,3.1} \\
    T4  & put the black bowl in the bottom drawer of the cabinet and close it                    & 86.7\,{\scriptsize$\pm$\,4.2}  & 96.0\,{\scriptsize$\pm$\,2.0}  & \textbf{100.0}\,{\scriptsize$\pm$\,0.0} \\
    T5  & put the white mug on the left plate and put the yellow and white mug on the right plate & 64.0\,{\scriptsize$\pm$\,5.3}  & \textbf{81.3}\,{\scriptsize$\pm$\,1.2} & 78.7\,{\scriptsize$\pm$\,1.2} \\
    T6  & pick up the book and place it in the back compartment of the caddy                     & 53.3\,{\scriptsize$\pm$\,8.1}  & \textbf{72.7}\,{\scriptsize$\pm$\,4.2} & 58.7\,{\scriptsize$\pm$\,1.2} \\
    T7  & put the white mug on the plate and put the chocolate pudding to the right of the plate & 63.3\,{\scriptsize$\pm$\,1.2}  & 78.7\,{\scriptsize$\pm$\,8.1}  & \textbf{84.7}\,{\scriptsize$\pm$\,9.9} \\
    T8  & put both the alphabet soup and the cream cheese box in the basket                      & 40.0\,{\scriptsize$\pm$\,10.4} & 62.0\,{\scriptsize$\pm$\,6.0}  & \textbf{84.0}\,{\scriptsize$\pm$\,2.0} \\
    T9  & put both moka pots on the stove                                                        & \textbf{3.3}\,{\scriptsize$\pm$\,3.1} & \textbf{3.3}\,{\scriptsize$\pm$\,3.1} & 1.3\,{\scriptsize$\pm$\,1.2} \\
    T10 & put the yellow and white mug in the microwave and close it                             & 38.7\,{\scriptsize$\pm$\,7.0}  & 63.3\,{\scriptsize$\pm$\,13.3} & \textbf{66.7}\,{\scriptsize$\pm$\,6.4} \\
    \midrule
    & Average                                                                                 & 50.5\,{\scriptsize$\pm$\,2.0}  & 64.8\,{\scriptsize$\pm$\,2.8}  & \textbf{72.1}\,{\scriptsize$\pm$\,1.1} \\
    \bottomrule
  \end{tabularx}
\end{table}

\subsection{Experimental Details}

\subsubsection{Base Policy} 

We adopt Diffusion Policy \cite{chi2023diffusion} and Rectified Flow Policy \cite{liu2022flow,zhang2025reinflow} as the base policies in \ours{} due to their strong ability to model complex robot action distributions. Both policies employ Transformer-based architectures, serving as the noise predictor and velocity predictor for diffusion and flow matching, respectively. Additionally, following \cite{chi2023diffusion}, we use two historical states as policy inputs and predict $T_p$ future actions, where $T_p = 16$ for Diffusion Policy and $T_p = 10$ for Rectified Flow Policy. We denote the predicted action chunk by $A_t=(\action_{t:t+T_p - 1})$.

\paragraph{Diffusion Policy.} During training, we follow the original Diffusion Policy implementation of \cite{chi2023diffusion}. At inference time, given a state $\state$, an initial action chunk $A^K$ is sampled from a standard Gaussian distribution. A learned denoising network $\epsilon_{\theta}(A^k, \state, k)$ then iteratively denoises $A^k$ over $K$ diffusion steps, yielding a sequence of intermediate action chunks and ultimately the final noise-free action chunk $A^0$. This process can be written as:
\begin{equation}
    A^{k-1} = \alpha_k(A^{k} - \beta_k\epsilon_{\theta}(A^k, \state, k) + \mathcal{N}(0,\sigma_k^2I)),
\end{equation}
where $\alpha_k, \beta_k, \sigma_k$ are functions of the iteration step $k$ that define the noise schedule. Here, the state $\state$ serves as the conditioning variable in the denoising process.

\paragraph{Flow Policy.} Flow matching \cite{liu2022flow} offers a streamlined and generalized alternative to traditional diffusion models by learning a continuous vector field that maps a simple, tractable prior to the target data distribution. We adopt the Rectified Flow policy used in \cite{zhang2025reinflow} in our work. During training, the learnable velocity field $v_{\theta}(\cdot)$ learns to transport Gaussian noise $A^0 \sim \mathcal{N}(0, I)$ to an expert action $A^1$ via the probability path $A^{\tau} = (1-\tau)A^0 + \tau A^1$, where $\tau \in [0, 1]$ is the flow matching step. The policy is trained by regressing $v_{\theta}(\cdot)$ to the target velocity $A^1 - A^0$ as follows:
\begin{equation}
    \mathcal{L}_{\text{FM}}(\theta) = \mathbb{E}_{\tau, A^0\sim\mathcal{N}(0, I), (A^1, \state)}\left[\|A^1 - A^0 - v_{\theta}(A^{\tau}, \tau, \state)\|_2^2\right].
\end{equation}
We define $\{\tau_k\}_{k=0}^K$ as a discretization of the time interval $[0, 1]$, with $\Delta \tau = \tau_{k+1} - \tau_k$. At inference, actions are generated by solving the ODE $\frac{dA^{\tau}}{d\tau} = v_{\theta}(A^{\tau}, \tau, \state)$ via the Euler update:
\begin{equation}
    A^{\tau_{k+1}} = A^{\tau_k} + \Delta\tau v_{\theta}(A^{\tau_k}, \tau_k, \state).
\end{equation}

\subsubsection{Contrastive Preference Learning for Generative Policies}

As our backbone preference learning algorithm, we adopt a variant of contrastive preference learning (CPL) \cite{hejna2024contrastive}, following \cite{miao2025uncertainty} to align the generative policies. Specifically, the training objective for the diffusion policy is defined as follows:
\begin{equation}\label{eq:cpl_dp_loss}
\begin{aligned}
    \mathcal{L}_{\text{DP}}^{(\sigma^w, \sigma^l)}(\theta) &= -\mathbb{E}_{
    \action_t^{k,w} \sim q(\action_t^{k,w}|\action_t^{0,w}), \
    \action_t^{k,l} \sim q(\action_t^{k,l}|\action_t^{0,l}), \
    k \sim \mathcal{U}(1, K)} \Bigg[
    \log \sigma \Bigg( -\alpha K \cdot \\
    &\quad \Bigg(
    \sum_{t=1}^L \gamma^t 
    \left(
    \|\epsilon_t^w - \epsilon_\theta(\action_t^{k,w}, \state_t^w, k)\|_2^2 - 
    \|\epsilon_t^w - \epsilon_{\text{ref}}(\action_t^{k,w}, \state_t^w, k)\|_2^2 
    \right) \\
    & \quad - \sum_{t=1}^L \gamma^t 
    \left( 
    \|\epsilon_t^l - \epsilon_\theta(\action_t^{k,l}, \state_t^l, k)\|_2^2 - 
    \|\epsilon_t^l - \epsilon_{\text{ref}}(\action_t^{k,l}, \state_t^l, k)\|_2^2 
    \right) 
    \Bigg) \Bigg)
    \Bigg],
\end{aligned}
\end{equation}
where $\sigma^w$ and $\sigma^l$ denote the preferred and rejected trajectory segments, respectively; $k$ is the diffusion timestep, $K$ is the number of diffusion denoising steps, $\action^{k, w} \sim q(\action_t^{k, w} | \action_t^{0, w})$ denotes the ground-truth action $\action^{0, w}$ corrupted with noise $\epsilon_t^w$ after $k$ forward diffusion steps. The noisy rejected action $\action^{k, l}$ is defined similarly.

Similarly, for the flow policy, we adapt the CPL objective by replacing the diffusion denoising error with the velocity-prediction error. In Rectified Flow, the network regresses the target velocity between the ground-truth action and the sampled noise. Thus, the preference learning objective for the flow policy is defined as:
\begin{equation}\label{eq:cpl_fp_loss}
\begin{aligned}
    \mathcal{L}_{\text{FP}}^{(\sigma^w, \sigma^l)}(\theta) &= -\mathbb{E}_{
    \action_t^{0,w}, \action_t^{0,l} \sim \mathcal{N}(0, \mathbf{I}),
    \tau \sim \mathcal{U}(0, 1)} \Bigg[ 
    \log \sigma \Bigg( -\alpha \cdot \\
    \Bigg( \sum_{t=1}^L \gamma^t &
    \left( 
    \|(\action_t^{1,w} - \action_t^{0,w}) - v_\theta(\action_t^{\tau,w}, \tau, \state_t^w)\|_2^2 - 
    \|(\action_t^{1,w} - \action_t^{0,w}) - v_{\text{ref}}(\action_t^{\tau,w}, \tau, \state_t^w)\|_2^2
    \right) \\
    - \sum_{t=1}^L \gamma^t &
    \left( 
    \|(\action_t^{1,l} - \action_t^{0,l}) - v_\theta(\action_t^{\tau,l}, \tau, \state_t^l)\|_2^2 - 
    \|(\action_t^{1,l} - \action_t^{0,l}) - v_{\text{ref}}(\action_t^{\tau,l}, \tau, \state_t^l)\|_2^2 
    \right) 
    \Bigg) \Bigg)
    \Bigg],
\end{aligned}
\end{equation}
where $\tau \in [0, 1]$ is the continuous flow-matching step, and $\action_t^{1, w}$ denotes the ground-truth preferred action. The term $\action_t^{0, w} \sim \mathcal{N}(0, \mathbf{I})$ is the initial sampled noise, and $\action_t^{\tau, w} = (1-\tau)\action_t^{0,w} + \tau \action_t^{1,w}$ represents the interpolated action along the probability path. The actions for the rejected segment $\action_t^{1, l}$, $\action_t^{0, l}$, and $\action_t^{\tau, l}$ are defined similarly.

In practice, because our generative policies predict action chunks of length $T_p$, evaluating the loss at every single timestep within a segment of length $L$ incurs prohibitive memory costs due to gradient accumulation. To alleviate this overhead and accelerate training, we stride the timestep summation in the learning objective by $T_p$. As a result, the loss is evaluated only at non-overlapping action chunks, substantially reducing memory consumption without compromising training stability. This strategy is applied to both diffusion and flow policies.

\subsubsection{Environmental Details}

We use Robomimic \cite{mandlekar2021matters}, a large-scale robotic manipulation benchmark designed for imitation learning and offline reinforcement learning (RL). It consists of five manipulation tasks, each accompanied by a proficient human (\textit{ph}) teleoperated demonstration dataset. Four tasks (excluding \emph{Tool Hang}) additionally provide mixed proficient/non-proficient human (\textit{mh}) demonstration datasets. In our experiments, we use the \textit{mh} datasets when applicable, as they are generally considered more challenging than the \textit{ph} datasets due to the inclusion of demonstrations from multiple non-proficient operators. The five tasks considered in this paper are:
\begin{itemize}
    \item \textit{Lift}: The robot arm must lift a small cube. 
    \item \textit{Can}: The robot must place a coke can from a large bin into a smaller target bin.
    \item \textit{Square}: The robot must pick a square nut and place it on a rod.
    \item \textit{Transport}: Two robot arms must transfer a hammer from a closed container on a shelf to a target bin on another shelf. One robot arm must retrieve the hammer from the container, while the other arm must clear the target bin by moving a piece of trash to the nearby receptacle. Finally, one arm must hand the hammer over to the other, which must place the hammer in the target bin.
    \item \textit{Tool Hang}: A robot arm must assemble a frame consisting of a base piece and hook piece by inserting the hook into the base, and hang a wrench on the hook. This is the most difficult task due to the multiple stages that each require precise, dexterous, rotation-heavy movements.
\end{itemize}

\subsubsection{Details of Baselines}
For both \textbf{Mixed BC} and \textbf{Filtered BC}, we fine-tune the pretrained policies using a small set of expert demonstrations combined with the exact same rollout trajectories used by \ours{}. For \textbf{Filtered BC}, any trajectories that fail to satisfy the task-specific success criteria are discarded. To ensure a fair comparison, the reinforcement learning baseline, \textbf{DPPO} \cite{ren2025diffusion}, is restricted to approximately the same number of environment interactions (rollouts) as our method. Furthermore, both $\textbf{CPL}_{KL}$ and \textbf{DPPO} utilize the same low-dimensional state inputs as all other methods. $\textbf{CPL}_{KL}$ relies on the ground-truth environment rewards provided by the benchmark (dense when available), whereas \textbf{DPPO} uses the sparse task rewards for policy optimization. All methods use low-dimensional state observations. For \ours{}, we additionally leverage the corresponding image observations (at a resolution of 224 $\times$ 224) solely for pseudo-preference label generation.

\subsubsection{Experimental Settings}\label{sec:detail_exp_setting}

For pseudo-preference label generation, we set the number of expert reference demonstrations to $M=10$ for \textit{Lift}, $M=20$ for \textit{Can}, and $M=40$ for the remaining tasks. In practice, we select the $M$ shortest demonstrations to form the reference set. To construct the initial preference dataset, we uniformly sample $N$ candidate segment pairs from the merged dataset—specifically, $N=10000$ for \textit{Lift}, $N=20000$ for \textit{Can}, and $N=30000$ for the remaining tasks—and perform a search over the discrimination threshold $\epsilon_{\sigma} \in \{0.3, 0.4, 0.6\}$. We select the highest threshold that successfully retains at least 1,000 valid preference pairs for \textit{Lift} and \textit{Can}, and 5,000 pairs for the remaining tasks. 

From a computational perspective, computing the ORCA metric in Eq.~(\ref{eq:orca_reward}) between a rollout and an expert demonstration incurs a dynamic programming cost of $\mathcal{O}(H \cdot \tilde{H})$ \cite{huey2025imitation}. To improve efficiency, the entire labeling process is performed offline prior to policy training. Specifically, instead of anchoring each rollout to a single demonstration, we compute and cache $M$ independent progress sequences per rollout trajectory, one for each expert demonstration. With these step-wise progress traces precomputed, evaluating the aggregated score for an arbitrary segment reduces to an $\mathcal{O}(M)$ operation. This requires only computing the marginal progress difference for each of the $M$ expert demonstrations and applying the max-pooling operation described previously. As a result, the proposed caching strategy eliminates redundant sequence-matching computations and substantially accelerates preference dataset construction. We follow the implementation of \cite{huey2025imitation} using cosine distance for the spatial cost matrix, while utilizing an adaptive temperature that scales dynamically with the minimum spatial cost to ensure robust temporal alignment.

Our implementation is built upon the codebases of \cite{chi2023diffusion} and \cite{miao2025uncertainty}. Additional hyperparameter details for \ours{} are provided in Table \ref{tab:shared_hyperparams_base_policy}.

\begin{table}[h]
\noindent
\caption{Comprehensive list of hyperparameters for \ours{}.}
\label{tab:shared_hyperparams_base_policy}
\centering
\begin{tabular}{ll}
\toprule
\textbf{Hyperparameter} & \textbf{Value} \\
\midrule
\multicolumn{2}{c}{\textit{Preference Generation (\ours{})}} \\
\midrule
Reference Demonstrations ($M$)       & 10-40 \\
Discrimination Threshold ($\epsilon_\sigma$) & $\{0.3, 0.4, 0.6\}$ \\
Visual Feature Encoder               & R3M \cite{nair2022r3m} \\
Distance Metric                      & Cosine Distance \\
Segment Length ($L$)                 & 100 (Lift, Can), 200 (Square), 300 (others) \\
\midrule
\multicolumn{2}{c}{\textit{Policy Optimization}} \\
\midrule
Optimizer                            & AdamW \\
Learning Rate                        & $1 \times 10^{-4}$ \\
Weight Decay                         & $1 \times 10^{-3}$ \\
Batch Size                           & 128 \\
Training Epochs                      & 1000 \\
Denoise Steps                        & 100 (Diffusion), 8 (Flow) \\
\bottomrule
\end{tabular}
\end{table}

\end{document}